\documentclass[a4paper, times, 10pt,twocolumn]{article}
\usepackage[top=4.9cm,bottom=3.7cm,left=1.5cm,right=1.5cm]{geometry}
\usepackage{ICMLC}
\usepackage{times}
\usepackage{graphicx}
\usepackage{indentfirst}
\usepackage{latexsym}
\usepackage[margin=8pt,font=footnotesize,labelfont=bf,labelsep=period
]{caption}
\usepackage{amsmath,amssymb,amsfonts}
\usepackage{algorithmic}
\usepackage{textcomp}
\usepackage{xcolor}
\usepackage{booktabs}

\newsavebox\CBox
\def\textBF#1{\sbox\CBox{#1}\resizebox{\wd\CBox}{\ht\CBox}{\textbf{#1}}}

\pdfpagewidth=\paperwidth
\pdfpageheight=\paperheight
\title{DEEP GENERATIVE MODELS AS AN ADVERSARIAL ATTACK STRATEGY FOR TABULAR MACHINE LEARNING}
\author{\bf{\normalsize{SALIJONA DYRMISHI$^1$, MIHAELA CĂTĂLINA STOIAN$^2$, ELEONORA GIUNCHIGLIA$^3$, MAXIME CORDY$^1$}}\\ % The author should be in Upper case
\\
\normalsize{$^1$University of Luxembourg, Luxembourg}\\
\normalsize{$^2$University of Oxford, United Kingdom} \\
\normalsize{$^3$Imperial College London, United Kingdom} \\
\\
\normalsize{E-MAIL: salijona.dyrmishi@uni.lu, mihaela.stoian@st-hildas.ox.ac.uk}\\
 \normalsize{\quad e.giunchiglia@imperial.ac.uk, maxime.cordy@uni.lu }
}

\begin{document}
\maketitle
\thispagestyle{empty}

\begin{abstract}
   {Deep Generative Models (DGMs) have found application in computer vision for generating adversarial examples to test the robustness of machine learning (ML) systems. Extending these adversarial techniques to tabular ML presents unique challenges due to the distinct nature of tabular data and the necessity to preserve domain constraints in adversarial examples. In this paper, we adapt four popular tabular DGMs into adversarial DGMs (AdvDGMs) and evaluate their effectiveness in generating realistic adversarial examples that conform to domain constraints.}
\end{abstract}
\\
%-------------------------------------------------------------------------

%-------------------------------------------------------------------------

\Section{Introduction}
Deep Generative Models (DGMs) generate synthetic data after learning the probability distribution of their training data. They are most commonly used to augment datasets for better predictive performance of machine learning (ML) models \cite{han2005imbalance},  to promote fairness \cite{breugel2021_fairness}, ensure privacy \cite{lee2021invertible} etc. Several works in computer vision have repurposed DGMs as a tool to generate adversarial examples for ML models. Such adversarial examples pose a significant security threat by minimally altering original inputs and forcing the models to wrongfully change their predictions. In this scenario, adversarial DGMs (AdvDGMs) take as an input an original example and output an adversarial one while trying to minimize the adversarial and perturbation loss, in addition to their original loss functions. AdvDGMs promise shorter generation times compared to today's popular iterative adversarial attacks, which is beneficial and important for adversarial hardening \cite{ijcai2018p543}. 

Beyond computer vision, extending AdvDGMs to test and improve the robustness of tabular ML models against adversarial examples is challenging due to the unique characteristics of tabular data. These models must account for diverse feature types and preprocessing while ensuring the generated adversarial examples adhere to domain-specific constraints. For instance, in a credit scoring system, the \textit{``average transaction amount''}  must not exceed the \textit{``maximum transaction amount.''}  Violating such constraints results in unrealistic examples that do not map to real-world transaction history. Current tabular DGMs often fail in this regard, producing up to 100\% unrealistic examples \cite{stoian2024how}. Few efforts have been made to adapt DGMs into AdvDGMs for tabular data \cite{zhao2021attackgan, usama2019generative, lin2022idsgan, hu2022generating}, however, these attempts often focus on a single use case, use generic models not tailored for tabular data, and handle the realism of their outputs by modifying only independent features, thus limiting the adversarial example search space.

In this paper, we convert four popular tabular DGMs into AdvDGMs and evaluate their potential to generate successful adversarial examples that fulfill three objectives: satisfy constraints, change model prediction, and maintain minimal distance from the original input. To boost the performance of tabular AdvDGMs, we extend them with a constraint repair layer \cite{stoian2024how}, ensuring that the outputs always satisfy domain constraints. Adding the constraint repair layer should not significantly impact the efficiency of AdvCDGMs compared to iterative adversarial techniques. Hence, we investigate the impact of CL on runtime. Finally, we compare our AdvDGMs' performance with three attacks from literature optimized for domain constraints. The source code, the data and the models are publicly available. \footnote{https://github.com/salijona/C-AdvDGM}.

\Section{Related Work}
\subsection{DGMs for Tabular Data}
Several approaches based on DGMs have been specifically designed to address particular challenges in generating tabular data such as mixed types of features, and imbalanced categorical data. Notable among these are GAN-based approaches like TableGAN \cite{park2018_tableGAN}, CTGAN \cite{xu2019_CTGAN}, OCT-GAN \cite{kim2021oct}, and IT-GAN \cite{lee2021invertible}. These methods leverage the power of GANs to model the underlying data distribution and generate synthetic samples that closely resemble real-world tabular data. Following privacy concerns, two approaches, i.e., DPGAN~\cite{xie2018differentially} and PATE-GAN~\cite{Jordon2019privacy}, incorporate differential privacy techniques to ensure that the generated synthetic data does not reveal sensitive information about the individuals in the original dataset. Alternativly to GANs, Xu et al. \cite{xu2019_CTGAN} proposed TVAE as a variation of the standard Variational AutoEncoder, while TabDDPM \cite{kotelnikov2023tabddpm} and STaSy \cite{kim2023stasy} were proposed following the achievements of score-based models. Finally, Liu et al. \cite{liu2022goggle} proposed GOGGLE, a model that uses graph learning to infer relational structure from the data.

\subsection{DGMs as an attack strategy}
\label{sec:related_AdvDGM}
In pioneering work by Xiao et al. \cite{ijcai2018p543}, Generative Adversarial Networks (GANs) were employed to generate adversarial examples for images. Their approach utilized the original examples as input to the generator, with the output representing the perturbation added to the original image. The generator was trained using a combination of adversarial loss, obtained by evaluating the perturbed instances on the target model, and GAN loss based on the discriminator's predictions. Subsequent research has aimed at refining the architectures and methodologies for generating adversarial examples with DGMs. For instance, Jandial et al.\cite{jandial2019advgan++} utilized the latent representations of images as input to the generator, considering them more prone to adversarial perturbations. Meanwhile, Bai et al.\cite{bai2021ai} introduced a novel attacker in the loop to train the discriminator adversarially. In another approach, Ding et al.\cite{ding2021advfoolgen} explored  VAE-GAN and concatenated original images with noise vectors as input to the generator. Moreover, Song et al.\cite{song2018constructing} focused on GAN-based adversarial examples not restricted by Lp norms. 

DGM-based adversarial methods have been applied to malware and intrusion detection tasks \cite{zhao2021attackgan, usama2019generative, lin2022idsgan, hu2022generating} using similar architectures as those for images. Challenges arise from ensuring the realism of generated examples in these new domains, with researchers addressing this by preserving non-functional features, thus reducing the search space for adversarial examples.

\section{Problem Statement}
\label{sec:iclr-problem}

\subsection{Deep Generative Models - DGM}
In standard generative modeling, we aim to learn parameters for a generative model ($p_{\theta}$) that approximates an unknown distribution $p_X$ based on a training dataset $\mathcal{D}$ of $N$ samples drawn from $p_X$. The DGM model can then output synthetic samples that closely follow the training data distribution. To exemplify, we describe the Generative Adversarial Network (GAN) as a common model architecture in the literature for synthesizing tabular data. A GAN consists of two neural networks: a generator $G$ and a discriminator $D$. The generator takes random noise as input and aims to generate synthetic data samples that are indistinguishable from real data, while the discriminator aims to distinguish between real and fake samples. Both $G$ and $D$ are trained iteratively in a min-max optimization task:

\vspace{-1em}
\begin{equation}
\begin{split}
\min_G \max_D V(D, G) = \mathbb{E}_{x \sim p_x}[\log D(x)] \\ + \mathbb{E}_{z \sim p_z}[\log(1 - D(G(z)))] 
\end{split}
\end{equation}

where $p_z$ is the noise distribution and $V$ is the function that the discriminator $D$ wants to maximize and generator $G$ wants to minimize.

\subsection{Adversarial Deep Generative Models - AdvDGM}
For AdvDGM, the input to the generator $G$ is no longer the noise vector but the initial point $x$, for which we aim to generate the adversarial example $\widetilde{x}=x + \delta$.  We introduce a target classifier $h$ for which we want to test the robustness using the examples generated by a DGM.  Additionally, let $y$ represent the target class label, and $x$ denote the input sample. The adversarial loss $\mathcal{L}_{adv}$ is computed as:

\vspace{-1em}
\begin{equation}
\mathcal{L}_{adv} = \max_{\| \delta \| \leq \epsilon} \left[ \ell(h(p_{\theta}(x + \delta)), y) \right]
\end{equation}

where $\delta$ denotes the perturbation added to the input sample $x$, constrained by its magnitude $\epsilon$ such that $| \delta | \leq \epsilon$, and $\ell$ denotes the loss function used for classification.

Moreover, the perturbation loss $\mathcal{L}_{pert}$ measures the magnitude of modifications required to transform a legitimate sample into an adversarial one. It is calculated as:

\vspace{-1em}
\begin{equation}
\mathcal{L}_{pert} = \| \delta \|
\end{equation}

where $\delta$ is the perturbation added to the input sample. 

The total loss of the AdvDGM model combines the initial loss of DGMs altogether, with adversarial and perturbation loss as follows:

\vspace{-1em}
\begin{equation}
\mathcal{L}_{AdvDGM} = \mathcal{L}_{DGM} - \alpha*\mathcal{L}_{adv} + \beta*\mathcal{L}_{pert}
\end{equation}

where $\alpha$ and $\beta$ are scaling factors and $\mathcal{L}_{DGM}$ is specific to the DGM used for modeling the data. For GANs specifically, $\mathcal{L}_{DGM}$ can be defined as:

\vspace{-1em}
\begin{equation}
\mathcal{L}_{DGM} = \frac{1}{m} \sum_{i=1}^{m} \log(1 - D(G(z^{(i)})))
\end{equation}

\subsection{Domain constraints}
The sample space of $p_X$ provides some knowledge on the acceptable values for each feature within its range but also in relationship with other features. Let $\Pi$ be a set of constraints expressing this background knowledge. We assume each constraint in $\Pi$ to be a linear inequality involving variables $x_k$ corresponding to features of the dataset. These inequalities take the form:

\vspace{-1em}
\begin{equation}
\sum_k w_kx_k + b \geq 0
\end{equation}

where $w_k$ are coefficients, $b$ is a constant, and $\geq$ denotes either greater than or equal to. A sample generated by a DGM assigns values to these variables. If the inequality is true for these assigned values, then the sample satisfies the constraints.

\section{Constrained Tabular Adversarial Deep Generative Models}

To ensure the adversarial examples generated by AdvDGMs comply with constraints \(\Pi\), we extend them to include the constraint repair layer $CL$ from Stoian et al. \cite{stoian2023}. The $CL$ layer takes as input i) domain constraints expressed as linear inequalities  ii) a feature repair ordering, and iii) an original example.  The example goes through an evaluation check for constraint satisfaction, and if any constraints are violated, the example will be minimally modified so that it is guaranteed that the resulting example will respect the constraints. This differentiable layer can be integrated during training, noted as C-AdvDGM, or used only during sampling, noted as P-AdvDGM.

\begin{figure}[ht!]
    \vspace{-0.1cm}
    \centering
    \includegraphics[width=0.8\linewidth]{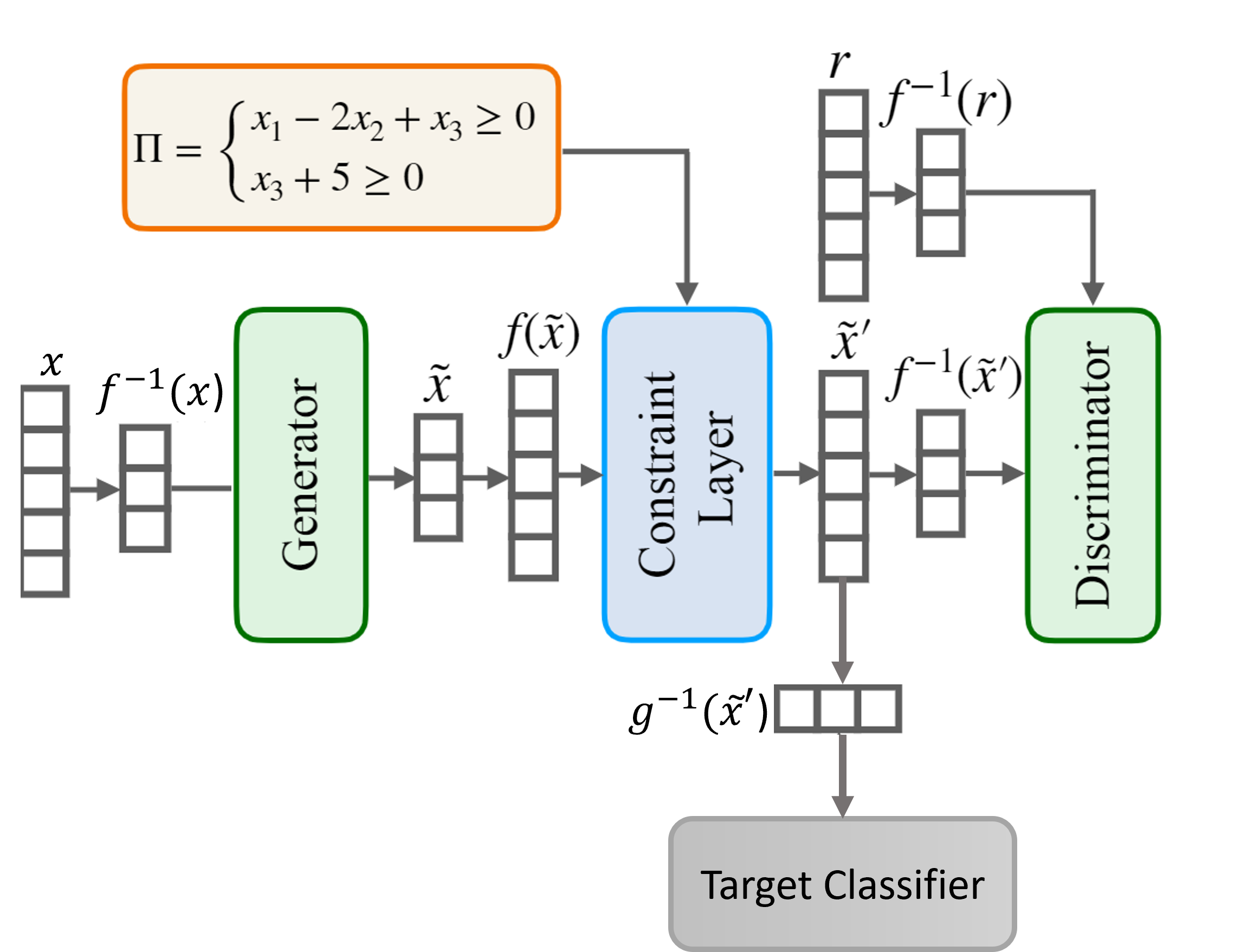}
    \caption{Overview of a C-AdvDGM based on GAN.}
    \vspace{-0.2cm}
    \label{fig:overview-advDGM}
\end{figure}

Figure \ref{fig:overview-advDGM} gives an overview of C-AdvDGMs for GANs, however the same can be applied to any tabular DGM. The generator takes as input the initial example $x$ transformed through a mapping function $f^{-1}$ (i.e min-max scaling) and outputs $\widetilde{x}$, which is transformed back into the original data space before undergoing constraint evaluation and repair via the constrained layer. The resulting constrained example $\widetilde{x}'$ is then transformed into the space used by the GAN before being fed into the discriminator to calculate $L_{GAN}$. Additionally, $\widetilde{x}'$ is transformed by a function $g$ into the space of the target classifier to compute $L_{adv}$. It's noteworthy that in some cases, $f = g$. When the constrained layer is operating during the training time, $f$ needs to be differentiable too.

\section{Experimental Settings}

\label{sec:exp_settings}

\begin{table*}[t!]
\centering
\small
\caption{ASR $\uparrow$ of tabular AdvDGMs ($\epsilon$ = 0.5). In bold the best success rate among AdvDGM, P-AdvDGM and C-AdvDGM for each model in case it is greater than the error rate of the models on original data.}
\label{tab:rob_accuracy}
\begin{tabular}{@{}lrrrr|rrrr|rrrr@{}}
\toprule
& \multicolumn{12}{c}{Target model}  \\ \cmidrule(lr){2-13} \
& \multicolumn{4}{c}{TorchRLN} & \multicolumn{4}{c}{VIME} & \multicolumn{4}{c}{TabTransformer}  \\ \cmidrule(lr){2-5} \cmidrule(lr){6-9} \cmidrule(lr){10-13}
Attack $\backslash$ Dataset & URL & WiDS & Heloc & FSP & URL & WiDS & Heloc & FSP & URL & WiDS & Heloc & FSP\\
\midrule
-                 & 0.04                          & 0.19                          & 0.28                          & 0.24                          & 0.06                            & 0.19                            & 0.26                             & 0.37                           & 0.09                             & 0.22                             & 0.28                           & 0.35                       \\ \cmidrule(r){1-1}
AdvWGAN           & \(\textBF{0.73}\) & 0.03          & 0.31          & 0.30          & 0.32          & 0.02          & 0.34         & 0.20          & 0.59          & 0.04          & 0.33          & 0.25          \\
P-AdvWGAN         & \(\textBF{0.73}\) & 0.07          & \(\textBF{0.93}\) & 0.70          & \textbf{0.34}          & 0.16       & \textbf{0.77}          & \textbf{0.50}          & \textbf{0.60}          & 0.22        & 0.73          & 0.45          \\
C-AdvWGAN         & 0.52          & 0.17 & 0.46          & \(\textBF{0.73}\) & 0.15          & 0.08          & 0.75          & \textbf{0.50}          & \textbf{0.60}          & 0.18          & \textbf{0.95}          & \textbf{0.54}          \\ \cmidrule(r){1-1}
AdvTableGAN       & \(\textBF{0.14}\) & 0.03          & 0.15          & 0.08          & 0.08          & 0.02          & 0.07          & 0.13          & 0.09          & 0.02          & 0.13          & 0.13          \\
P-AdvTableGAN     & \(\textBF{0.14}\) & 0.17 & 0.28 & \(\textBF{0.28}\) & 0.08          & 0.12                        & 0.18          & \textbf{0.38}          & 0.09                & 0.20                  & 0.22          & 0.47                                \\
C-AdvTableGAN     & 0.09          & 0.12          & 0.09          & 0.27          & 0.08          & 0.12          & 0.12          & \textbf{0.38}          & 0.09          & 0.14          & 0.27          & 0.43          \\ \cmidrule(r){1-1}
AdvCTGAN          & 0.01          & 0.01          & 0.18          & 0.02          & 0.04          & 0.02          & 0.22          & 0.14          & 0.07          & 0.03          & 0.18          & 0.07          \\
P-AdvCTGAN        & 0.01          & 0.19 & 0.28          & 0.06          & 0.04                           & 0.18                             & 0.26                             & 0.37                             & 0.07                             & 0.22                            & 0.26                            & \textbf{0.46}                          \\
C-AdvCTGAN        & 0.02 & 0.16          & \(\textBF{0.37}\) & \(\textBF{0.32}\) & 0.04          & 0.14          & \textbf{0.27}          & 0.37          & 0.07          & 0.20          & \textbf{0.32}          & \textbf{0.46}          \\ \cmidrule(r){1-1}
AdvTVAE           & 0.00          & 0.00          & 0.18          & 0.06          & 0.01          & 0.00                             & 0.16          & 0.09          & 0.01         & 0.00                             & 0.17          & 0.13          \\
P-AdvTVAE         & 0.00          & 0.12 & 0.32          & 0.23          & 0.01          & 0.11                          & \textbf{0.28}          & 0.35          & 0.01                          & 0.14                            & 0.28          & \textbf{0.42}          \\
C-AdvTVAE         & 0.01 & 0.10          & \(\textBF{0.60}\) & \(\textBF{0.28}\) & 0.01          & 0.10                            & 0.27          & 0.37          & 0.01         & 0.12                             & 0.28          & \textbf{0.43}          \\ \bottomrule
\end{tabular}
\end{table*}

% underwent augmentation with our $\CL$, leading to the creation of C-WGAN, C-TableGAN, C-CTGAN, C-TVAE, and C-GOGGLE models.  Further, we turn four of these models into C-AdvWGAN, C-AdvTableGAN, C-AdvCTGAN, C-AdvTVAE, and C-AdvGOGGLE. For more details on tabular DGMs architecture, please refer to Appendix \ref{app:models}.

% \subsection{Datasets}
\textbf{Datasets.} We used four real-world datasets (\textit{URL}, \textit{WiDS}, \textit{Heloc}, \textit{FSP}) with domain constraints identified in literature \cite{stoian2024how}. 

% \subsection{Target models}

\textbf{Target models.} We used three tabular neural network classifiers (TorchRLN, VIME, TabTransformer) from \cite{simonetto2023constrained} for which we performed a hyperparameter search on each dataset to obtain the best parameters.

% \subsection{AdvDGMs} 
\textbf{Tabular DGMs.} Our experimentation involved four distinct tabular DGMs: WGAN \cite{Arjovsky2017_WGAN}, TableGAN \cite{park2018_tableGAN}, CTGAN \cite{xu2019_CTGAN}, TVAE \cite{xu2019_CTGAN}. 
Each model was updated according to the steps in 3.2 to obtain AdvWGAN, AdvTableGAN, AdvCTGAN, AdvTVAE. The modifications included as well discarding conditional loss for CTGAN, and the label classifier for TableGAN.  Then $CL$ was added to obtain the P-AdvDGMs and C-AdvDGMs versions of these models. The layer was extended to support, in addition to linear equalities, constraints of type ``if - else'' as conjunctions. 
We performed a hyperparameter search independently for AdvDGMs and C-AdvDGMs, exploring values of $\alpha$ and $\beta$ in the range of $[1, 100]$ and learning rate equal to $\{0.001, 0.005, 0.01, 0.05\}$. We used the random variable ordering as an input to the constrained layer CL. 

\textbf{SOTA attacks.} To compare the performance of our tabular AdvDGMs We used two gradient attacks CPGD and CAPGD and a genetic algorithm attack MOEVA \cite{simonetto2022, simonetto2023constrained}. 

\textbf{Metrics.} We measured Attack Success Rate (ASR) as the ratio of the adversarial examples that have $\epsilon <0.05$ ($L_2$ norm), cause the model to change its prediction and satisfy the constraints, -- over the total number of original examples.  All the metrics are reported as average over 5 runs.

\section{Results}
\label{sec:experiments}

\subsection{Adversarial generation capability}
\label{sec:adv_gen}

Table \ref{tab:rob_accuracy} shows the ASR of our AdvDGMs and their constrained counterparts for the four datasets under study. 

The results demonstrate that for all target models, only AdvWGAN and its constrained counterparts are successful in significantly increasing the error rate of the model by reaching an ASR of up to 95\% for Heloc dataset with TabTransformer model. All the AdvDGMs and their constrained counterparts are unsuccessful on WiDS dataset, having an ASR lower than the error rate of the models on original non-adversarial data. 

% The other models manage to increase the error rate only in maximum of 2 dataset and at most to 60\% (C-AdvTVAE for Heloc on TorchRLN). 

Regarding the addition of the constrained layer $CL$, the results show that it is beneficial in increasing the ASR of the attacks. Out of 48 cases, P-AdvDGMs have higher ASR than AdvDGMs 38 times with a maximum difference of 62\% (P-AdvGAN on Heloc and TorchRLN). Similarly C-AdvDGMS have higher ASR 37 times with a maximum difference of 62\% (C-AdvGAN on Heloc and TabTransformer).

\begin{table}[ht!]
\centering
\footnotesize
\setlength{\tabcolsep}{2.5pt}
\caption{Training time and sample generation time in seconds for adversarial models.}
\label{tab:runtime}
\begin{tabular}{@{}lllll|llll@{}}
\toprule
& \multicolumn{4}{c}{Training Time (min)} & \multicolumn{4}{c}{Sample Time (s)} \\ \cmidrule(lr){2-5} \cmidrule(lr){6-9}
& URL & WiDS & Heloc & FSP & URL & WiDS & Heloc & FSP \\ \midrule
AdvWGAN & 2.25 & 5.17 & 3.88 & 1.17 & 0.02 & 0.10 & 0.00 & 0.00 \\
C-AdvWGAN & 5.32 & 24.53 & 6.53 & 2.33 & 0.03 & 0.16 & 0.00 & 0.01 \\ \cmidrule(r){1-1}
AdvTableGAN & 1.07 & 47.38 & 7.47 & 1.08 & 0.31 & 5.51 & 0.05 & 0.08 \\
C-AdvTableGAN & 1.13 & 48.78 & 7.63 & 1.12 & 0.33 & 5.63 & 0.06 & 0.08 \\ \cmidrule(r){1-1}
AdvCTGAN & 2.00 & 19.65 & 3.35 & 0.78 & 2.09 & 29.59 & 0.16 & 0.26 \\
C-AdvCTGAN & 2.78 & 28.83 & 6.88 & 1.10 & 2.08 & 29.32 & 0.21 & 0.26 \\ \cmidrule(r){1-1}
AdvTVAE & 1.05 & 9.33 & 1.22 & 0.40 & 2.08 & 29.41 & 0.16 & 0.26 \\
C-AdvTVAE & 2.03 & 20.67 & 2.38 & 0.78 & 2.09 & 29.20 & 0.16 & 0.26 \\ \bottomrule
\end{tabular}
\end{table}

\begin{table*}[ht!]
\centering
\small
\caption{ASR $\uparrow$ of three existing attacks in literature, and the best ASR for AdvWGAN and its constrained counterparts. The best values are in bold and second best are underlined.}
\label{tab:sota-comparison}
\begin{tabular}{@{}lrrrr|rrrr|rrrr@{}}
\toprule
& \multicolumn{12}{c}{Target model}  \\ \cmidrule(lr){2-13} 
& \multicolumn{4}{c}{TorchRLN} & \multicolumn{4}{c}{VIME} & \multicolumn{4}{c}{TabTransformer}  \\ \cmidrule(lr){2-5} \cmidrule(lr){6-9} \cmidrule(lr){10-13}
Attack $\backslash$ Dataset& URL & WiDS & Heloc & FSP & URL & WiDS & Heloc & FSP & URL & WiDS & Heloc & FSP \\ \midrule
- & 0.04  & 0.19  & 0.28 & 0.24  & 0.06  & 0.19 & 0.26 & 0.37 & 0.09  & 0.22  & 0.28  & 0.35 \\ 
 CPGD  & \textbf{0.94} & \textbf{0.45} & 0.27 & 0.23 & \textbf{0.62} & \textbf{0.47} & 0.31 & 0.36 & \textbf{0.85} & \underline{0.61} & 0.27 & 0.45 \\ 
CAPGD  & 0.83 & 0.41 & 0.32 & 0.25 & \underline{0.59} & \textbf{0.43} & 0.30 & 0.36 & 0.59 & 0.53 & 0.38 & 0.46 \\ 
MOEVA  & \underline{0.86} & \textbf{0.45} & \textbf{0.99} & \textbf{0.92} & 0.55 & 0.41 & \textbf{0.99} & \textbf{0.74} & \textbf{0.85} & \textbf{0.64} & \textbf{1.00} & \textbf{0.74} \\ 
% CAA    & 0.95 & 0.51 & 0.99 & 0.92 & 0.67 & 0.53 & 0.99 & 0.77 & 0.88 & 0.77 & 1 & 0.74 \\ 
        \midrule
*-AdvWGAN & 0.73 & 0.17 & \underline{0.93} & \underline{0.73} & 0.34&0.16 & \underline{0.77} & \underline{0.50} & 0.60& 0.22& \underline{0.95} & \underline{0.54} \\ \bottomrule
\end{tabular}
\end{table*}

\subsection{Constrained Layer impact on runtime}\label{sec:exp_time}

\textBF{Train time:}
The results in Table \ref{tab:runtime}  show that the constrained models require at most 4.7 more time to train compared to the unconstrained model (C-AdvWGAN for WiDS). On the other hand, for some models, the constrained and unconstrained versions take the same time to train as in the case of TableGAN.

\textBF{Sampling time:} 

From the results in Table \ref{tab:runtime}, we observe that C-AdvDGMs exhibit, at most, a 0.12-second increase in runtime compared to their unconstrained counterparts (notably, C-AdvTableGAN for the WiDS dataset). On average, the runtime is 0.02 seconds slower for C-AdvWGAN and 0.04 seconds slower for C-AdvTableGAN. Contrarily, it is 0.06 seconds faster for C-CTGAN and 0.05 seconds faster for C-AdvTVAE. This indicates that our constrained layer incurs negligible overhead, especially when used at sampling time, enabling AdvCDGMs to remain viable for practical applications.

\subsection{Comparison with SOTA}

Table \ref{tab:sota-comparison} demonstrates that  our best performing attack *-AdvWGAN ranks as the second-best attack for Heloc and FSP datasets on all three target models. On these dataset, the gradient attacks CPGD and CAPGD perform poorly with a maximum increase of the model's error rate of 11\%. The genetic attack MOEVA has the highest success rate in 9 out of 12 cases.

\section{Conclusion}

DGMs are efficient adversarial attack methods in computer vision, but adapting them for tabular data poses challenges due to the properties of tabular data and current tabular DGMs not respecting domain constraints. In this paper, we adapted four tabular DGMs into AdvDGMs and extended them into C(P)-AdvDGMs by adding a constraint repair layer. Notably, only AdvWGAN consistently achieved high success rates in both unconstrained and constrained versions, which is surprising given that WGAN is older and not always the most performant in dataset augmentation literature. Our experiments showed that including the constraint layer during training or sampling improves the success rate of AdvDGMs, highlighting the importance of compliance with background knowledge for adversarial attacks. Further investigation is needed to understand why the constraint layer is more successful in training vs. sampling.

%-------------------------------------------------------------------------
\section*{Acknowledgements}
Mihaela Cătălina Stoian was supported by the EPSRC under the grant EP/T517811/1. Salijona Dyrmishi was supported by the Luxembourg National Research Funds (FNR) AFR Grant 14585105.

%-------------------------------------------------------------------------

\bibliographystyle{IEEEtran}
\bibliography{references_iclr}

\end{document}